\documentclass[journal]{IEEEtran}
\usepackage{times}
\usepackage{amssymb}
\usepackage{helvet}
\usepackage{mathrsfs}
\usepackage{url}
\usepackage{courier}
\usepackage{booktabs,caption}
\usepackage{threeparttable}
\usepackage{graphicx}
\usepackage{float} 

\usepackage{overpic}
\usepackage{color}
\usepackage[english]{babel}
\usepackage[T1]{fontenc}
\usepackage[utf8]{inputenc}
\usepackage{multirow}
\usepackage{diagbox}
\usepackage{authblk}
\usepackage[numbers, compress]{natbib}
\usepackage{soul}
\usepackage{xcolor}
\usepackage{footnote}
\usepackage{hyperref}

\usepackage{makecell}

\usepackage{bbm}
\usepackage{xspace}
\usepackage{subfig}
\usepackage{amsmath}
\usepackage{algorithm,algorithmic}
\usepackage{wrapfig}
\usepackage{lipsum}
\usepackage{verbatim}

\makeatletter
\DeclareRobustCommand\onedot{\futurelet\@let@token\@onedot}
\def\@onedot{\ifx\@let@token.\else.\null\fi\xspace}
 
\def\ie{\emph{i.e}\onedot}

\def\etal{\emph{et al}\onedot}
\makeatother

\begin{document}

\title{\tnnls{Dynamic} Sub-Cluster-Aware Network for Few-Shot Skin Disease Classification}

\author{Shuhan Li, Xiaomeng Li,
	Xiaowei Xu and Kwang-Ting Cheng

\thanks{S. Li is with the Department of Computer Science and Engineering, The Hong Kong University of Science and Technology, Hong Kong, SAR.  X. Xu is with Guangdong Provincial People's Hospital, Guangdong Academy of Medical Sciences, Guangzhou, China. X. Li and K. Cheng are with the Department of Electronic and Computer Engineering, The Hong Kong University of Science and Technology, Hong Kong, SAR, China. (e-mail: eexmli@ust.hk; timcheng@ust.hk).  
(Corresponding author: Xiaomeng Li.)

{Copyright (c) 2020 IEEE. Personal use of this material is permitted. Permission from IEEE must be obtained for all other uses, including reprinting/republishing this material for advertising or promotional purposes, collecting new collected works for resale or redistribution to servers or lists, or reuse of any copyrighted component of this work in other works.}}}

%

%


\newcommand{\para}[1]{\vspace{.05in}\noindent\textbf{#1}}

\newcommand{\xmli}[1]{{\color[rgb]{0.9,0.2,0.9}{[XM: #1]}}}

\newcommand{\revise}[1]{{\color[rgb]{0,0,0}{#1}}}

\newcommand{\tnnls}[1]{{\color[rgb]{0,0,0}{#1}}}

\newcommand{\tnnlsminor}[1]{{\color[rgb]{0,0,0}{#1}}}

\definecolor{mycolor}{RGB}{0,0,0} 

\maketitle



\IEEEpeerreviewmaketitle

\begin{abstract}
This paper addresses the problem of few-shot skin disease classification by introducing a novel approach called the \textbf{S}ub-\textbf{C}luster-\textbf{A}ware \textbf{N}etwork (SCAN) that enhances accuracy in diagnosing rare skin diseases. \tnnlsminor{The key insight motivating the design of SCAN is the observation that skin disease images within a class often exhibit multiple sub-clusters, characterized by distinct variations in appearance. To improve the performance of few-shot learning, we focus on learning a high-quality feature encoder that captures the unique sub-clustered representations within each disease class, enabling better characterization of feature distributions.}
Specifically, SCAN follows a dual-branch framework, where the first branch learns class-wise features {\bf \emph{to distinguish different skin diseases}}, and the second branch aims to learn features which can effectively partition each class into several groups so as {\bf \emph{to preserve the sub-clustered structure within each class}}. 
To achieve the objective of the second branch, we present a cluster loss to learn image similarities via unsupervised clustering. To ensure that the samples in each sub-cluster are from the same class, we further design a purity loss to refine the unsupervised clustering results.
We evaluate the proposed approach on two public datasets for few-shot skin disease classification. The experimental results validate that our framework outperforms the state-of-the-art methods by around 2\% to 5\% in terms of sensitivity, specificity, accuracy, and F1-score on the SD-198 and Derm7pt datasets. 


\if 1 
The initial diagnosis on clinical images of skin lesion diseases is essential to the patients due to the convenience of collecting images as well as the efficient procedures compared to a dermatological examination by pathologists. 
However, training a reliable deep network model usually needs a large amount of data which is not sufficient for some rare skin conditions. Moreover, there exist large intra-class variations in skin diseases, especially in clinical images, caused by different body parts, illumination and shooting distances, etc. 
This paper designs a novel multi-task few-shot learning framework for skin disease classification with a low data regime and large intra-class diversity to improve the predictions on rare skin conditions. 
Built on the transfer-based few-shot learning method, we introduce xxx which combines the supervised classification with online deep clustering to further discover finer groups within each labeled class. We experimentally show that the learned features can reflect the similarity relationships for each class and adapt more discriminative features on unseen classes.
We evaluate the proposed approach on two public clinical image datasets for skin lesion disease classification. The experimental results validate that our framework outperforms the baseline by a large margin and surpasses other few-shot learning methods. 
\fi 

\end{abstract}

\section{Introduction}

\begin{figure}
\centering
\includegraphics[width=0.45\textwidth]{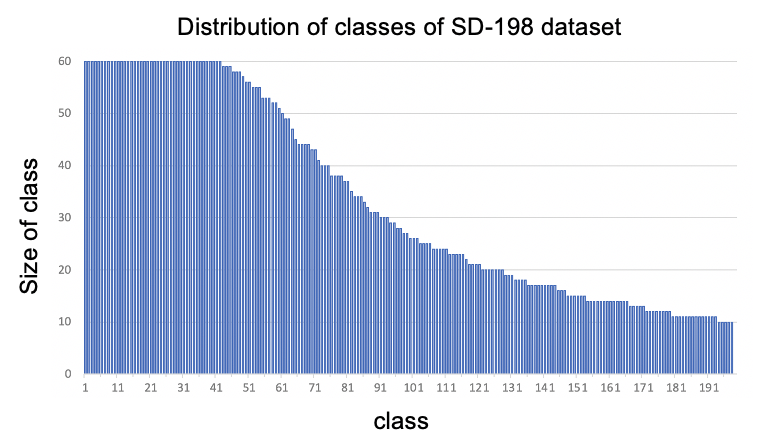} 
\caption{The long-tailed class distribution in the SD-198 dataset. The X-axis refers to 198 skin disease classes, and the Y-axis refers to the number of images in the corresponding class.} 
\vspace{-4mm}
\label{fig1} 
\end{figure}

In the past decade, advances in deep convolutional neural networks and the availability of a large number of annotated images have continued to push the boundaries in a variety of medical image analysis tasks, such as organ segmentation~\cite{yu2019uncertainty,li20183d}, tumor segmentation~\cite{cciccek20163d,li2018h,isensee2021nnu,li2020transformation} and disease screening~\cite{wang2017chestx,sun2021explanation,feng2020deep,chen2020transfer,li2019canet}.
Apart from the relatively sufficient examples of the common diseases, there are more than 6,000 known rare diseases, at much lower prevalence, affecting 7\% of the population worldwide~\cite{chung2021rare}. 
The diagnosis of rare conditions is challenging for clinicians due to a lack of experience and for machines due to a lack of clinical samples. 
Rare skin disease diagnosis is one of the application domains that suffer from data deficiency. 
For example, the SD-198 dataset~\cite{sun2016benchmark} consists of 6,584 skin images with 198 disease classes, and 70 of these classes contain relatively fewer images than the other skin disease classes, as shown in Figure~\ref{fig1}.
This paper aims to develop a deep learning model which is trained with images of common skin diseases while generalizing well on rare skin diseases.

\begin{figure}
\centering
\includegraphics[width=0.45\textwidth]{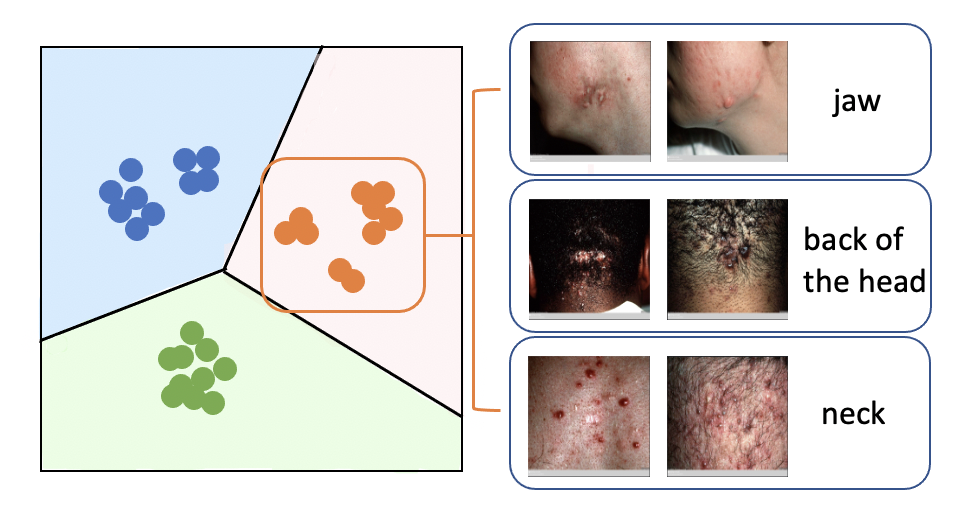} 
\caption{ Left: An example of features learned by our framework. Three classes are separated by their class labels (shown in different colors), and various sub-groups are generated automatically for each class. Right: An example of a class forming three intrinsic sub-groups, with different appearances in different body parts. } 
\vspace{-4mm}
\label{fig2} 
\end{figure}

To address the challenge of rare skin disease diagnosis, few-shot learning (FSL) has emerged as a promising solution. FSL aims to learn transferable knowledge on common classes (base classes) and apply it to the rare ones (novel classes) with only a handful of labeled data. 
Current FSL methods for few-shot skin disease classification can be broadly divided into two groups: meta-learning-based~\cite{prabhu2019few,li2020difficulty,mahajan2020meta,zhang2020st,zhu2020alleviating,zhu2021temperature,singh2021metamed,zhou2022few} and transfer-learning-based~\cite{guo2020broader,chen2021self,medina2020self,phoo2020self,dai2023pfemed,xiao2023boosting}. Meta-learning-based methods are mainly based on MAML~\cite{finn2017model} and Prototypical Networks~\cite{snell2017prototypical}, with a focus on different problems in rare skin disease classification. 
For example, Li~\etal~\cite{li2020difficulty} built a difficulty-aware meta-learning model based on MAML~\cite{finn2017model} to address the different weights of sampled tasks in the imbalanced skin dataset.
The other group of FSL methods, transfer-learning-based methods, aim to learn a powerful feature extractor in the \textit{pre-training} stage with sufficient data, which can be quickly adapted to novel classes in the \textit{fine-tuning} stage with a few novel examples. 
For example, Xiao~\etal~\cite{xiao2023boosting} designed a multitask framework that leverages contrastive learning as the auxiliary task to improve the performance of the classification task.


\tnnls{However, there exists a key challenge for rare skin disease classification that has rarely been addressed by the previous FSL methods. Skin disease images have large intra-class variations~\cite{anand2022skin} and form multiple latent groups within a class, \ie, sub-clusters, for some classes, as shown in Figure~\ref{fig2}. To address this issue, the Prototypical Clustering Networks (PCN)~\cite{prabhu2019few} was proposed to learn multiple prototypes for each class. In their experiments, they manually set a fixed number of clusters for all classes, typically dividing each class into 2 sub-clusters. This approach may lead to inaccurate structure learning for certain classes, as different classes often display varying sub-cluster structures. Instead of a fixed sub-cluster number for all classes, we should identify various sub-cluster numbers for different classes.}

Therefore, in this paper, we design a novel dual-branch \textbf{S}ub-\textbf{C}luster-\textbf{A}ware \textbf{N}etwork, named SCAN, for few-shot skin disease classification. \tnnlsminor{Our approach stands out from previous methods by dynamically identifying the unique sub-clustered structures within each class. This inovation is inspired by the discovery of intrinsic sub-clustered structures in skin disease datasets.
By learning these structures from skin images, the network is able to extract more precise feature representations on the base classes, leading to enhanced performance on novel classes.}
\tnnlsminor{To achieve this objective, SCAN incorporates a powerful feature encoder that consists of two branches. The class branch learns class-wise features {\bf \emph{to distinguish different skin diseases}}, while the cluster branch learns sub-clustered features {\bf \emph{to preserve the sub-clustered structure within each class}}.}
\tnnlsminor{To facilitate the learning of dynamic sub-clustered representation, we introduce a cluster branch with a cluster loss, which employs unsupervised clustering to partition all base class samples into several clusters.} To ensure that the samples in each sub-cluster are from the same class, we further design a purity loss to refine the unsupervised clustering results. \tnnlsminor{Inspired by the concept of triplet loss, the purity loss guides the anchor away from the wrong cluster center and pulls it closer to the correct class center.}
Notably, our framework can {identify the different numbers of sub-clusters according to the different class structures}.
In the fine-tuning stage, we evaluate the performance of the pre-trained encoder on novel classes with sampled episodes. 

Experimental results on two skin disease benchmark datasets, SD-198~\cite{sun2016benchmark} and Derm7pt~\cite{kawahara2018seven}, show that our method outperforms \tnnlsminor{the state-of-the-art methods in both 1-shot and 5-shot settings. Our method achieves significant improvements compared to the prior best method in various metrics such as sensitivity, specificity, accuracy, and F1-score, with approximately 2\% improvement observed for the SD-198 dataset and 5\% improvement for the Derm7pt dataset.}
In summary, our contributions can be summarized as follows:
\begin{itemize}
	\item  \tnnls{We present a dual-branch framework for rare skin disease classification following the paradigm of transfer-learning-based FSL methods. Our framework effectively identifies inherent and unique sub-cluster structures within each common disease class, leading to improved classification accuracy for rare disease classes.}
        
	\item \tnnls{Within our framework, we develop a powerful feature encoder that combines supervised classification (class branch) and unsupervised clustering (cluster branch). Moreover, we introduce the purity loss to improve the result of intra-class clustering in the unsupervised cluster branch.}

	\item Our method is evaluated on two public skin disease datasets, SD-198~\cite{sun2016benchmark} and Derm7pt~\cite{kawahara2018seven}, and achieves state-of-the-art results for few-shot skin disease classification. Code is available at \href{https://github.com/xmed-lab/SCAN}{https://github.com/xmed-lab/SCAN}.
	
\end{itemize}

\begin{figure*} [!t]
\centering
\includegraphics[width=1.00\textwidth]{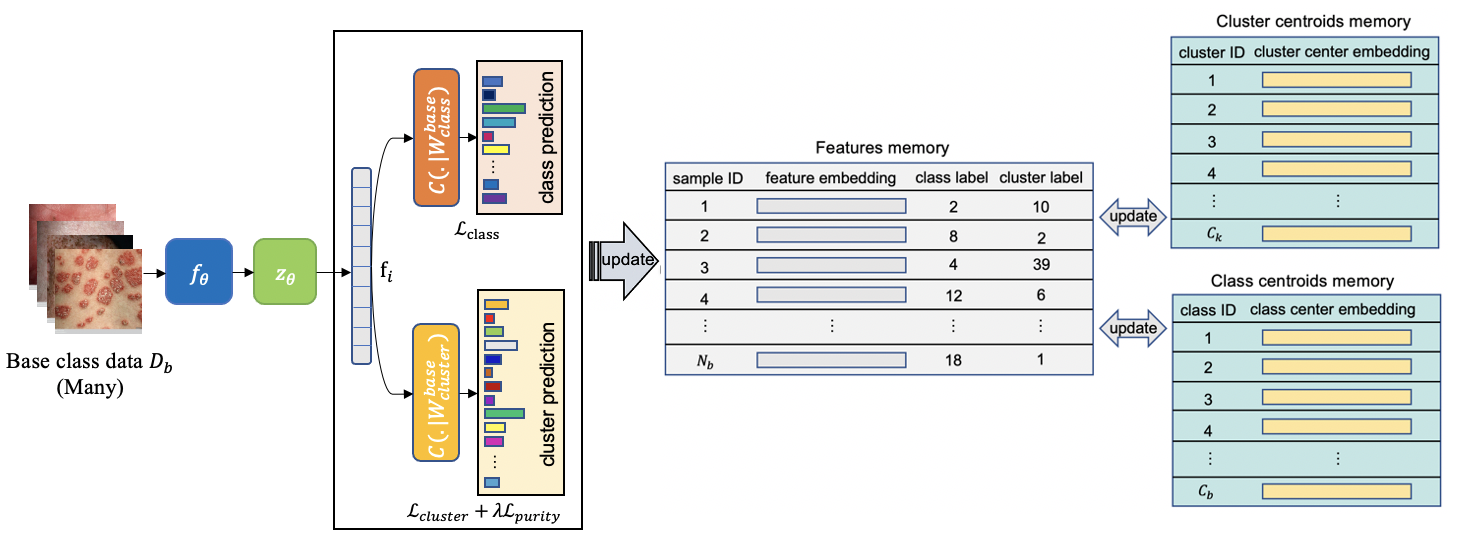} 
\caption{Overview of proposed Sub-Cluster-Aware Network (SCAN). We train this network on base class data in pre-training stage.} 
\vspace{-4mm}
\label{fig3} 
\end{figure*}

\begin{figure*} [!t]
\centering
\includegraphics[width=1.00\textwidth]{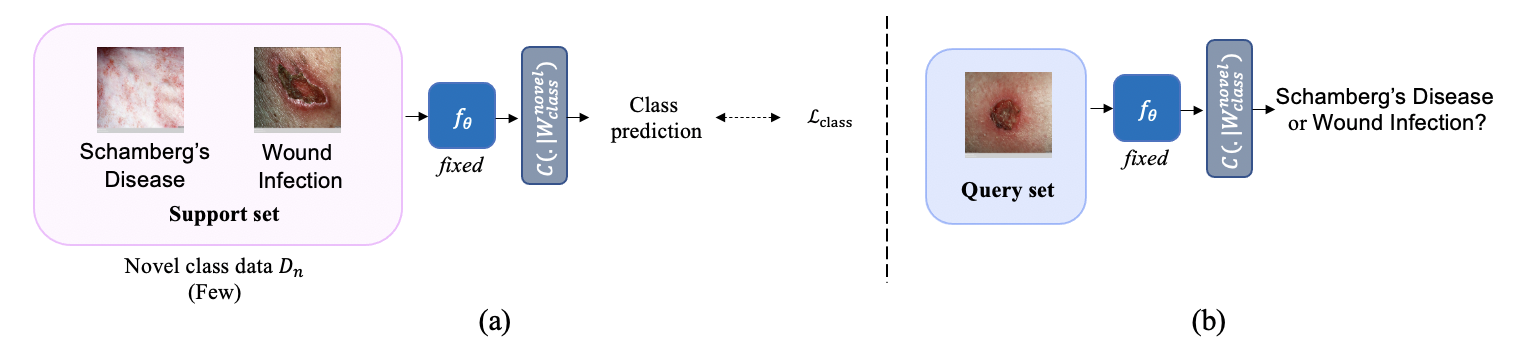} 
\caption{An example of fine-tuning stage for a 2-way 1-shot task on novel class set. We utilize the feature encoder $f_{\theta}$ trained in the pre-training stage and fix the parameters. (a): In each episode, a classifier is trained by the support set images. (b): Evaluating on the query set images.
} 
\vspace{-4mm}
\label{fig4} 
\end{figure*}

\section{Related Work}

\subsection{Few-shot Learning in Computer Vision}
Few-shot learning (FSL) aims to acquire prior knowledge from base classes (with large labeled examples) and propagate it to novel classes (with insufficient labeled examples, unseen during the training stage). FSL approaches can be categorized into two groups based on their training processes: meta-learning-based and transfer-learning-based methods.

Meta-learning-based methods, which are motivated by meta-learning principles, have been extensively explored in FSL~\cite{finn2017model,nichol2018first,vinyals2016matching,snell2017prototypical,li2020revisiting,liu2023bilaterally,lai2020learning,an2023dual,zhang2023semi}. Early works such as MAML \cite{finn2017model}, Reptile \cite{nichol2018first}, Matching Networks \cite{vinyals2016matching}, Prototypical Networks \cite{snell2017prototypical}, and Relation Networks \cite{li2020revisiting} adopt a meta-learning framework that trains the model episodically, enabling it to learn how to update the network using a few sampled examples. This enables these methods to achieve promising accuracy on novel classes with very limited labeled data. Meta-learning-based methods can be further divided into two sub-branches. The first sub-branch is gradient-based methods~\cite{}, such as MAML and Reptile, which aim to learn good initialization parameters for each episode. These methods can quickly adapt to unseen classes within one or a few gradient descent steps. The second sub-branch is metric-based methods, which utilize encoded feature vectors and employ a distance metric to assign labels based on the nearest-neighbor principle. These methods use similarity measures like cosine similarity (e.g., Matching Networks) or Euclidean distance (e.g., Prototypical Networks). 

In recent years, a second group of FSL methods based on transfer learning has emerged~\cite{chen2019closer,dhillon2019baseline,tian2020rethinking,wu2018improving,mangla2020charting,liu2020negative,yang2021free,hu2021leveraging,hu2022squeezing,bendou2022easy}. These methods, including Baseline \cite{chen2019closer}, S2M2\_R \cite{mangla2020charting}, and PT-MAP \cite{hu2021leveraging}, train a standard classification network on base classes and fine-tune the classifier head on episodes generated from novel classes. As learning meaningful feature representations is essential to various deep-learning tasks~\cite{chang2021comprehensive,zhang2022tn,li2022video,yan2021zeronas}, the transfer-learning-based methods aim to train a strong feature extractor that generates transferable features for the novel set. Experimental results have shown that simple transfer-learning-based methods, such as training a classification network with standard cross-entropy loss, can achieve comparable performance to previous FSL methods while offering a simpler and more effective process \cite{chen2019closer, wang2019simpleshot}. Due to their superior performance of transfer-learning-based methods, we explore them as ways to predict rare skin diseases.

\begin{figure*}
\centering
\includegraphics[width=1.00\textwidth]{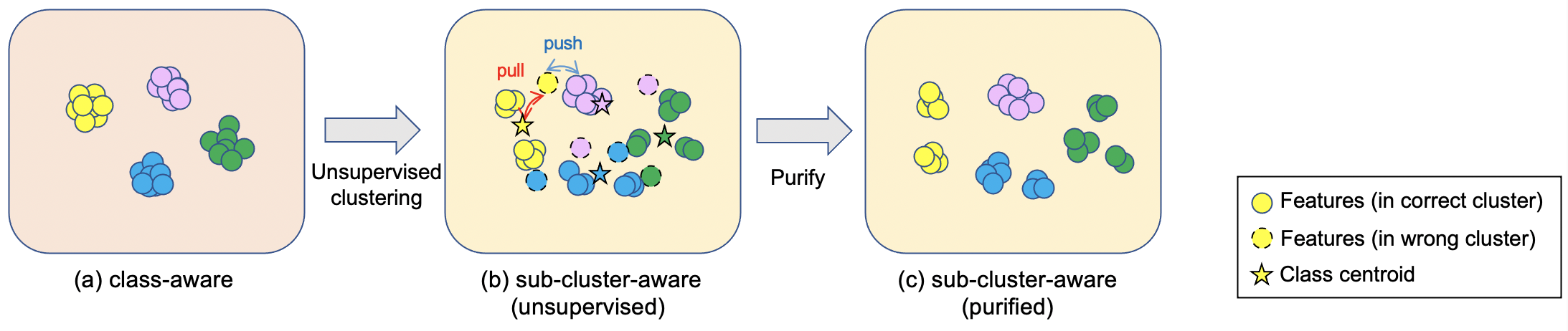} 
\caption{Effects of cluster loss $\mathcal{L}_{cluster}$ (from (a) to (b)) and purity loss $\mathcal{L}_{purity}$ (from (b) to (c)).} 
\vspace{-4mm}
\label{fig5} 
\end{figure*}

\subsection{Few-shot Learning for Skin Disease Classification}
Due to the unbalanced distribution of skin disease classes and the deficiency of images for rare conditions, it is essential to apply FSL methods to alleviate the reliance on massive training data. Previous studies have explored meta-learning-based methods, particularly within the gradient-based and metric-based sub-branches. In the gradient-based sub-branch, Li \etal \cite{li2020difficulty} built a Difficulty-Aware Meta-Learning method (DAML) method, which is based on MAML framework. DAML takes into account the varying difficulties of different episodic tasks. The model adjusts the weights of task-specific losses by down-weighting the easy tasks and increasing the weights of the hard tasks. This approach reduces the importance of easier tasks and emphasizes the significance of more challenging tasks. Singh et al. \cite{singh2021metamed} trained the Reptile~\cite{nichol2018first} model with additional regularization techniques such as mixup, cutout, and cutmix. These techniques enhance model generalization by introducing variations and augmentations during training. The incorporation of these regularization methods improved the model's performance by 2\%$\sim$5\%.
In the metric-based sub-branch, more methods are proposed. 
Mahajan~\etal~\cite{mahajan2020meta} replaced the conventional convolutional layers in Reptile and Prototypical Networks with group equivariant convolutions~\cite{cohen2016group}. This modification aims to capture invariant features after undergoing various transformations. Skin diseases typically lack a dominant global orientation structure, thus making group equivariant convolutions suitable for extracting informative features. Zhu \etal \cite{zhu2020alleviating} identified an incompatibility issue between cross-entropy loss and episode training and proposed the query-relative loss. This novel loss function further enhances the behavior of metric-based methods. Additionally, Zhu \etal \cite{zhu2021temperature} extended the Prototypical Networks framework by introducing different temperatures for different categories. By penalizing query samples that are not sufficiently close to their corresponding categories, this approach strengthens the generalization ability of the learned metric.
For the category of transfer-learning-based algorithms, Xiao~\etal~\cite{xiao2023boosting} designed a multitask framework that leverages contrastive learning as the auxiliary task to enhance the performance of the few-shot classification task. Dai~\etal~\cite{dai2023pfemed} proposed a dual-encoder architecture to combine the knowledge from a large-scale image dataset and a few-shot medical image dataset. Some other works~\cite{chen2021self,medina2020self,phoo2020self} utilized unlabeled data and self-supervised learning to mitigate the performance degradation of few-shot learners. In our work, we do not use additional data from other domains or unlabeled data. We focus on training the effective feature encoder on common skin disease images and transferring that knowledge to rare skin disease images.


Additionally, few methods consider the sub-cluster structures within skin disease classes, except for Prototypical Clustering Networks (PCN) proposed by Prabhu \etal \cite{prabhu2019few}. The authors extend Prototypical Networks by representing each class as multiple prototypes, instead of a single one. However, PCN divides each class into fixed $k$ sub-clusters, which is determined without considering the specific clustering results for different classes. In contrast to PCN, our designed framework aims to dynamically learn the specific intrinsic sub-clusters for each skin disease type. By doing so, we can generate more accurate feature representations for transference to unseen classes. Our approach goes beyond the fixed $k$ sub-cluster division in PCN and allows for a more flexible and tailored representation learning process.

\section{Methodology}
\subsection{Problem Setting}

In few-shot learning, data are partitioned into two disjoint sets: the base set and the novel set. We denote the images from the base set as $D_b=\{(x_i,y_i)\}^{N_b}_{i=1}$, which contains $N_b$ labeled images from $C_b$ base classes and $y_i \in \{1,...,C_b\}$.
Let $D_n=\{(x_i,y_i)\}^{N_n}_{i=1}$ denote the images from the novel set, with few samples from $C_n$ novel classes, where $C_b\cap C_n=\emptyset$. 
The main goal of few-shot skin disease classification is to learn rich and transferable feature representations from the base set, such that they can easily be adapted to the novel set with only a few labeled data.

To evaluate the generalization ability on the unseen classes, we sample \textit{N-way K-shot} episodes from novel set $D_n$ following the most common way proposed by Vinyals~\etal~\cite{vinyals2016matching}. Each episode contains a \textit{support set} for fast adaptation and a \textit{query set} for model evaluation. In the support set, $N$ classes are randomly selected from $C_n$ novel classes and $K$ labeled images are sampled from each class, denoted as $S=\{(x_i, y_i)\}^{N\times K}_{i=1}$. In the query set, $q$ images are picked from the same $N$ classes without their labels, denoted as $Q=\{(x_i)\}^{N\times q}_{i=1}$. The label information of query set $Q$ is only available when computing the accuracy.

Our work is based on the transfer-learning FSL paradigm. Therefore, we train a feature encoder in the pre-training stage with the data from the base set, \revise{as shown in Figure \mbox{\ref{fig3}}}. Then, we preserve the feature extractor $f_{\theta}$ of the trained model and freeze the network parameters. In the following fine-tuning stage, we fine-tune a new classifier by using the images from the support set in a sampled episode, and evaluate the accuracy on the query set, \revise{as shown in Figure \mbox{\ref{fig4}}}.

\subsection{Framework Overview}

To specifically identify the sub-clusters within each class, we design the Sub-Cluster-Aware Network (SCAN) in the pre-training stage. The overall framework is shown in Figure \ref{fig3}, which consists of a dual-branch network structure as well as several auxiliary memory banks.

The dual-branch network includes a feature extractor $f_\theta$ for feature encoding, a projection head $z_\theta$ for dimensional reduction, and two linear classifiers $C(.|\mathbf{W}_{class}^{base})$, $C(.|\mathbf{W}_{cluster}^{base})$ for class and cluster prediction, respectively. We train the network using the labeled data in the base set by minimizing three losses: $\mathcal{L}_{class}$, $\mathcal{L}_{cluster}$ and $\mathcal{L}_{purity}$. $\mathcal{L}_{class}$ is a standard cross-entropy loss computed for the class branch, which aims to classify skin disease according to the class labels. The other two losses are used in the cluster branch, with more details introduced in section~\ref{method-C}.

\subsection{Sub-Cluster-Aware Network}
\label{method-C}
Due to the intrinsic sub-cluster structure in the skin disease datasets, the model trained with the class-wise cross-entropy loss cannot learn the structural relationships \textbf{\textbf{within}} each skin disease class, as shown in {Figure \ref{fig5}(a)}. 
Considering the sub-cluster structures for skin disease images, we present a novel Sub-Cluster-Aware Network (SCAN) to learn sub-cluster embeddings for skin disease classification.

To achieve this, we create three memory banks, \ie, feature memory, class centroid memory, and cluster centroid memory; as shown in Figure~\ref{fig3}. \tnnls{The feature memory bank saves the feature embeddings and their class and cluster labels. The size of the feature memory bank is set as $N_b$, which is the total number of training images. At the beginning of the training, we utilize the feature extractor $f_{\theta}$ to embed all the training images into feature embeddings and save them in the feature memory bank. Then, we initialize the cluster labels by applying K-means on all the saved feature embeddings.}
 The class centroid memory saves the center embedding of each class by averaging all the features from the corresponding class. The size of the class centroid memory bank is set to the total number of classes in the base class dataset. Similar to the class centroid memory, the cluster centroid memory stores the mean feature of each cluster, \revise{and the size is the number of total clusters}.

Then, in each forward propagation, the network and memory banks are updated through the procedures in the following sections.

\subsubsection{Forward propagation}
We derive the feature embedding ${\rm \bf f}_i = z_{\theta}(f_{\theta}(x_i))\in \mathbb{R}^d$ for an input image $x_i$, where $d$ refers to the dimension of ${\rm \bf f}_i$. After sending ${\rm \bf f}_i$ to two classifiers, we obtain the class probability prediction $p_i\in \mathbb{R}^{C_b}$ and the cluster probability prediction $p_i^{\prime}\in \mathbb{R}^{C_k}$, correspondingly. Here, ${C_b}$ is the number of base classes and ${C_k}$ is the number of clusters. We set $C_k > C_b$ such that the model can acquire finer structures than the original granularity defined by the class labels. 

\subsubsection{Computing class loss \texorpdfstring{$\mathcal{L}_{class}$}{Lcla}}
A standard cross-entropy loss in Equation~\ref{ClassCELoss} is used to compute class loss {$\mathcal{L}_{class}$}:
\begin{equation}
\mathcal{L}_{class} = \frac{1}{N_b} \sum_{i=1}^{N_b} \sum_{c=1}^{C_b} - y_{i,c}\log p_{i,c}, \label{ClassCELoss} 
\end{equation}  
\begin{equation}
y_{i,c}=\mathbbm{1}[y_i=c], \label{ClassIndicator} 
\end{equation}  
where $N_b$ denotes the amount of training data in base classes, and $\mathbbm{1}(\cdot)$ is the indicator function. $y_{i,c}$ is equal to 1 if $c$ is the correct class label of input image $x_i$, otherwise, $y_{i,c}$ is 0. $p_{i,c}$ denotes the $c_{th}$ item in the class probability vector $p_i$ which represents the probability of $x_i$ predicted as class $c$.

\subsubsection{Computing cluster loss \texorpdfstring{$\mathcal{L}_{cluster}$}{Lclu}}
With the cluster label of feature ${\rm \bf f}_i$ recorded in the feature memory, denoted as $y_i^{\prime}$, we calculate the cluster loss $\mathcal{L}_{cluster}$ by Equation~\ref{ClusterCELoss}:
\begin{equation}
\mathcal{L}_{cluster}= \frac{1}{N_b} \sum_{i=1}^{N_b} \sum_{s=1}^{C_k} -y_{i,s}^{\prime}\log p_{i,s}^{\prime}.\label{ClusterCELoss}
\end{equation}
\begin{equation}
y_{i,s}^{\prime}=\mathbbm{1}[y_i^{\prime}=s], \label{ClusterIndicator}
\end{equation}

Similarly, $y_{i,s}^{\prime}$ is equal to 1 if $s$ is the cluster label of input image $x_i$, otherwise $y_{i,s}^{\prime}$ is 0. $p_{i,s}^{\prime}$ denotes the cluster probability of $x_i$ predicted as cluster $s$.

\subsubsection{Computing purity loss \texorpdfstring{$\mathcal{L}_{purity}$}{Lp}}
The above two losses, $\mathcal{L}_{class}$ and $\mathcal{L}_{cluster}$, are effective in forming the sub-cluster structure of the features. However, the clustering may not precisely occur within the class as the class labels are not utilized during cluster prediction. Samples from different classes are likely to be wrongly gathered into the same cluster because of the inter-class similarity, as shown in {Figure \ref{fig5}(b)}. To address this issue, we propose a purity loss $\mathcal{L}_{purity}$ (Equation~\ref{ClusterPurityLoss}) to refine the clustering results:

\begin{equation}
\begin{aligned}
\mathcal{L}_{purity} & = \sum_{i=1}^{N_b} \mathcal{L}_{triplet}({\rm \bf f}_i, {\rm \bf f}_i^p, {\rm \bf f}_i^n) \\
& = \sum_{i=1}^{N_b} max(\|{\rm \bf f}_i-{\rm \bf f}_i^p\|_2^2 - \|{\rm \bf f}_i - {\rm \bf f}_i^n\|_2^2 + \alpha, 0), \label{ClusterPurityLoss}
\end{aligned}
\end{equation}
where ${\rm \bf f}_i$ is the anchor, the positive sample ${\rm \bf f}_i^p$ is the class center of the anchor's class, and the negative sample ${\rm \bf f}_i^n$ is the nearest feature in the same cluster of ${\rm \bf f}_i$ with a different class label. The margin between positive and negative pairs is denoted as $\alpha$. The purity loss is defined as the sum of the triplet loss over all the samples in the base set.

The purity loss refines the clustering results by pulling the anchor towards its correct class and pushing it away from the wrong cluster, as depicted in Figure \ref{fig5}(c). Specifically, the loss is activated only when the distance between the anchor and positive item is longer than the distance between the anchor and negative sample by a margin $\alpha$. By minimizing the purity loss, we can improve the quality of the learned sub-clusters and correct the improper clustering results caused by inter-class similarity.


\begin{table*} [t]
    \centering
     \caption{2-way 1-shot and 2-way 5-shot few-shot classification results (\%) on the SD-198 dataset. Results of other methods are produced by our re-implementation. $^{\dag}$Results reported in the paper. ``-'' refers to not reported in the paper.}
     
    \begin{tabular}{c|c|cccc}
    \toprule
    \multirow{2}{*}{Method} & \multirow{2}{*}{Backbone} & \multicolumn{2}{c}{2-way 1-shot} & \multicolumn{2}{c}{2-way 5-shot}   \\
    &  & Accuracy & F1-score & Accuracy & F1-score \\
    
    \midrule
        
         PCN~\cite{prabhu2019few} & \multirow{2}*{Conv4} & 70.03$\pm$1.42 &  70.78$\pm$1.61 & 84.95$\pm$1.15 & 85.87$\pm$1.12 \\
         \textbf{SCAN (ours)} & & {77.12$\pm$1.44}  & 78.00$\pm$1.51 & {90.22$\pm$0.95} & 91.01$\pm$0.90 \\
         \hline
         Meta-derm~\cite{mahajan2020meta} & \multirow{2}*{Conv6} & 65.3$^{\dag}$ & - & 83.7$^{\dag}$ & - \\
        \textbf{SCAN (ours)} & & {76.75$\pm$1.42}  & 77.64$\pm$1.50 & {87.45$\pm$1.08} & 88.28$\pm$1.03 \\
        \hline
         NCA~\cite{wu2018improving} & \multirow{8}*{WRN-28-10} & 71.27$\pm$1.50 & 71.27$\pm$1.50 & 83.30$\pm$1.20 & 84.23$\pm$1.19 \\
        ~ Baseline~\cite{chen2019closer} & & 75.72$\pm$1.47 & 76.64$\pm$1.56 & 88.95$\pm$1.00 & 89.66$\pm$0.97 \\
        ~ S2M2\_R~\cite{mangla2020charting} & & 76.42$\pm$1.52 & 77.51$\pm$1.59 & 90.32$\pm$0.89 & 90.97$\pm$0.89 \\
        ~ NegMargin~\cite{liu2020negative} & & 76.85$\pm$1.39 & 77.98$\pm$1.45 & 89.92$\pm$0.96 & 90.65$\pm$0.92 \\
        ~  \textcolor{mycolor}{PT+NCM~\cite{hu2021leveraging}} & & \textcolor{mycolor}{78.25$\pm$1.47} & \textcolor{mycolor}{78.86$\pm$1.47} & \textcolor{mycolor}{90.33$\pm$0.95} & \textcolor{mycolor}{90.90$\pm$0.93}\\
        ~ \textcolor{mycolor}{PEM$_{b}$E$\text{-}$NCM~\cite{hu2022squeezing}} & & \textcolor{mycolor}{78.32$\pm$1.48} & \textcolor{mycolor}{78.70$\pm$1.49} & \textcolor{mycolor}{90.48$\pm$0.96} & \textcolor{mycolor}{90.94$\pm$0.95}\\
        ~ \textcolor{mycolor}{EASY~\cite{bendou2022easy}} & & \textcolor{mycolor}{78.80$\pm$1.50} & \textcolor{mycolor}{79.44$\pm$1.51} & \textcolor{mycolor}{90.87$\pm$0.98} & \textcolor{mycolor}{91.43$\pm$0.96} \\
       \textbf{SCAN (ours)} & & \textbf{80.20$\pm$1.44} & \textbf{81.21$\pm$1.46} & \textbf{91.48$\pm$0.88} & \textbf{92.08$\pm$0.85}\\

    \bottomrule
    \end{tabular}
    \label{table:SD_similar_2way}
\end{table*}

\begin{table*} [t]
    \centering
     \caption{5-way 1-shot and 5-way 5-shot few-shot classification results (\%) on the SD-198 dataset. Results of other methods are produced by our re-implementation. $^{\dag}$Results reported in the paper. ``-'' refers to not reported in the paper.}
     
    \resizebox{\textwidth}{!}{
    \begin{tabular}{c|c|cccccccc}
    
    \toprule
    \multirow{2}{*}{Method} & \multirow{2}{*}{Backbone} & \multicolumn{4}{c}{5-way 1-shot} & \multicolumn{4}{c}{5-way 5-shot}   \\
    &  & Sensitivity & Specificity & Accuracy & F1-score & Sensitivity & Specificity & Accuracy & F1-score \\
    
    \midrule
        
         PCN~\cite{prabhu2019few} & \multirow{2}*{Conv4} & 44.79$\pm$0.96 & 86.20$\pm$0.24 & 77.92$\pm$0.38 & 45.59$\pm$1.03 &  64.78$\pm$0.99 & 91.19$\pm$0.25 & 85.91$\pm$0.39 & 65.70$\pm$1.02 \\
         \textbf{SCAN (ours)} & & 54.63$\pm$1.02 & 88.66$\pm$0.26 & 81.85$\pm$0.41 & 55.60$\pm$1.07 &  74.47$\pm$0.85 & 93.62$\pm$0.21 & 89.79$\pm$0.34 & 75.65$\pm$0.87 \\
         \hline
         Meta-derm~\cite{mahajan2020meta} & \multirow{2}*{Conv6} & - & - & - & - & - & - & - & -\\
        \textbf{SCAN (ours)} & & 53.43$\pm$1.06 & 88.36$\pm$0.27 & 81.37$\pm$0.42 & 54.07$\pm$1.24 & 73.83$\pm$0.90 & 93.46$\pm$0.22 &  89.53$\pm$0.36 & 74.73$\pm$0.92 \\
        \hline
         NCA~\cite{wu2018improving} & \multirow{8}*{WRN-28-10} & 45.43$\pm$1.02 & 86.36$\pm$0.26 & 78.17$\pm$0.41 & 45.91$\pm$1.08 & 61.83$\pm$0.96 & 90.46$\pm$0.24 & 84.73$\pm$0.38 & 62.83$\pm$1.01 \\
        ~ Baseline~\cite{chen2019closer} & & 51.57$\pm$1.05 & 87.89$\pm$0.26 & 80.63$\pm$0.42 & 52.54$\pm$1.11 & 73.65$\pm$0.94 & 93.41$\pm$0.24 & 89.46$\pm$0.38 & 74.71$\pm$0.96 \\
        ~ S2M2\_R~\cite{mangla2020charting} & & 54.79$\pm$1.07 & 88.70$\pm$0.27 &  81.91$\pm$0.43 & 55.49$\pm$1.13 & 77.11$\pm$0.83 & 94.28$\pm$0.21 & 90.84$\pm$0.33 & 78.17$\pm$0.84 \\
        ~ NegMargin~\cite{liu2020negative} & & 55.38$\pm$1.07 & 88.34$\pm$0.27 & 82.15$\pm$0.43 & 56.04$\pm$1.14 & 76.71$\pm$0.87 & 94.18$\pm$0.22 & 90.68$\pm$0.35 & 77.75$\pm$0.87 \\
        ~ \textcolor{mycolor}{PT+NCM~\cite{hu2021leveraging}} & & \textcolor{mycolor}{56.75$\pm$1.05} & \textcolor{mycolor}{89.19$\pm$0.26} & \textcolor{mycolor}{82.70$\pm$0.42} & \textcolor{mycolor}{56.91$\pm$1.11} & \textcolor{mycolor}{77.05$\pm$0.85} & \textcolor{mycolor}{94.01$\pm$0.21} & \textcolor{mycolor}{90.82$\pm$0.34} & \textcolor{mycolor}{78.12$\pm$0.88}\\
        ~ \textcolor{mycolor}{PEM$_{b}$E$\text{-}$NCM~\cite{hu2022squeezing}} & & \textcolor{mycolor}{57.26$\pm$1.06} & \textcolor{mycolor}{89.31$\pm$0.26} & \textcolor{mycolor}{82.90$\pm$0.42} & \textcolor{mycolor}{57.42$\pm$1.11} & \textcolor{mycolor}{77.99$\pm$0.89} & \textcolor{mycolor}{94.50$\pm$0.22} & \textcolor{mycolor}{91.43$\pm$0.36} & \textcolor{mycolor}{78.78$\pm$0.90} \\
        ~ \textcolor{mycolor}{EASY~\cite{bendou2022easy}} & & \textcolor{mycolor}{57.55$\pm$1.05} & \textcolor{mycolor}{89.39$\pm$0.26} & \textcolor{mycolor}{82.99$\pm$0.43} & \textcolor{mycolor}{57.77$\pm$1.12} & \textcolor{mycolor}{78.81$\pm$0.88} & \textcolor{mycolor}{94.71$\pm$0.21} & \textcolor{mycolor}{91.65$\pm$0.35} & \textcolor{mycolor}{79.53$\pm$0.89}\\
       \textbf{SCAN (ours)} & & \textbf{58.08$\pm$1.09} & \textbf{89.52$\pm$0.27} & \textbf{83.23$\pm$0.44} &
       \textbf{58.75$\pm$1.14} & \textbf{80.41$\pm$0.78} & \textbf{95.10$\pm$0.19} &
       \textbf{92.16$\pm$0.31} & \textbf{81.43$\pm$0.77} \\

    \bottomrule
    \end{tabular}
    }
    \vspace{-2mm}
    \label{table:SD_similar_5way}
\end{table*}

\subsubsection{Updating memory banks}
Finally, we update the contents in three memory banks according to the feature embeddings in the current epoch for the usage of next iteration.

In the feature memory bank, we smoothly update feature embeddings with a momentum decay rate $\beta \in (0,1]$:
\begin{equation}
{\rm \bf f}_{m,i} \leftarrow \beta \frac{{\rm \bf f}_i}{\left\|{\rm \bf f}_i\right\|} + (1-\beta) {\rm \bf f}_{m,i}, \label{MemoryUpdate}
\end{equation}
where ${{\rm \bf f}_i}$ is the feature embedding with sample ID $i \in \{1,2,...,N_b\}$, and ${\rm \bf f}_{m,i}$ represents the feature saved at the $ith$ location in the feature memory bank. The cluster label is assigned to item $i$ according to the nearest cluster center. The class labels remain the same since the ground truth does not change during training. 
In the class centroid memory bank, the centers of all the classes $c_i\in {C_b}$ are recalculated by the updated embeddings in the feature memory bank. 
In the cluster centroid memory bank, the centers of all the cluster $s_i\in {C_k}$ are recalculated by the updated embeddings and cluster labels.

To sum up, the whole loss function (Equation~\ref{WholeLossFunc}) of our proposed SCAN framework is the summation of three parts: 
\begin{equation}
\mathcal{L}_{SCAN} = \mathcal{L}_{class} + \mathcal{L}_{cluster} + \lambda \mathcal{L}_{purity}, \label{WholeLossFunc}
\end{equation}
where $\lambda$ is the weight for purity loss.

\section{Experiments}

\subsection{Datasets}
\textbf{SD-198 dataset}~\cite{sun2016benchmark} consists of 6,584 clinical images captured by mobile phones or digital cameras. The dataset collects a diverse range of 198 skin disease classes, with each class containing 10 to 60 examples. The images differ in color, exposure, and illumination, and contain a wide range of patients of various ages, genders, skin colors, disease locations, and durations. To fairly compare with existing methods, \tnnls{we follow the settings of previous work \cite{mahajan2020meta}. In their study, a subset of images is selected from 90 out of 198 classes, and these images are subsequently divided into base and novel sets. The base set consists of 20 classes, with each class containing 60 images. The novel set consists of 70 classes, with the number of images per class ranging from 10 to 19. In total, the base set contains 1200 images and the novel sst contains 965 images.}

\textbf{Derm7pt dataset}~\cite{kawahara2018seven} includes more than 2,000 color images from 20 skin lesion classes. 
In line with previous research \cite{mahajan2020meta}, we exclude two classes, namely ``miscellaneous'' and ``melanoma'', from our analysis. Subsequently, the remaining 18 classes are divided into base and novel sets. \tnnls{According to their settings, the base set consists of 13 classes, with the number of images per class varying from 40 to 698. The novel set consists of 5 classes, with each class containing 10 to 34 images. The base set contains a total of 1892 images, while the novel set contains 114 images.}

We resize all the input images to 80px$\times$80px for the above two datasets and apply the standard augmentations including random cropping, color jittering, rotation (-30 degrees to +30 degrees), and horizontal flipping.


\subsection{Implementation Details}
To fairly compare with existing few-shot methods, we employ \textbf{Conv4}, \textbf{Conv6} \cite{snell2017prototypical}, and \textbf{Wide ResNet (WRN)} \cite{zagoruyko2016wide} as backbones. To show the flexibility of our method, we further integrate our SCAN framework with \textbf{ResNet18} and \textbf{ResNet34}~\cite{he2016deep}.
For the projection head $z_{\theta}$, we use a non-linear structure $\left[ \rm fc\text{-}bn\text{-}relu\text{-}dropout\text{-}fc\text{-}relu\right ]$, which reduces the feature embedding dimensions to 256. Both classifiers $C(.|\mathbf{W}_{class}^{base})$ and $C(.|\mathbf{W}_{cluster}^{base})$ have the same structure, \ie,  a fully connected layer followed by a Softmax function. 
The number of classes $C_b$ is 20 for SD-198 and 13 for Derm7pt while the number of cluster $C_k$ is set to 40 and 25, respectively. For the hyperparameters in loss functions, we set $\alpha$ in Equation~\ref{ClusterPurityLoss} as 0.3, $\beta$ in Equation~\ref{MemoryUpdate} as 0.5, and $\lambda$ in Equation~\ref{WholeLossFunc} as 1. To train our SCAN model, we set the batch size to 16 and 64 for the SD-198 and Derm7pt datasets, respectively. We choose the SGD optimizer with a 0.0075 learning rate, 0.9 momentum and $10^{-5}$ weight decay for 800 epochs training. 


\begin{table*} [!htbp]
    \centering
    \caption{2-way 1-shot and 2-way 5-shot few-shot classification results (\%) on the Derm7pt dataset. Results of other methods are produced by our re-implementation. $^{\dag}$Results reported in the paper.}
    \begin{tabular}{c|c|cccc}
    \toprule
    \multirow{2}{*}{Method} & \multirow{2}{*}{Backbone} & \multicolumn{2}{c}{2-way 1-shot} & \multicolumn{2}{c}{2-way 5-shot}   \\
    &  & Accuracy & F1-score & Accuracy & F1-score \\
    \midrule
        
         PCN~\cite{prabhu2019few} & \multirow{2}*{Conv4} & 59.98$\pm$1.28 &  58.54$\pm$1.63 & 70.62$\pm$1.38 & 71.85$\pm$1.48 \\
        ~ \textbf{SCAN (ours)} & &{61.42$\pm$1.49} & 61.90$\pm$1.66 &  {72.58$\pm$1.28} & 74.05$\pm$1.32 \\
         \hline
         Meta-derm~\cite{mahajan2020meta} & \multirow{2}*{Conv6} & 61.8$^{\dag}$ & - &  76.9$^{\dag}$ & - \\
        ~ \textbf{SCAN (ours)} & & {62.80$\pm$1.34} & 63.75$\pm$1.50 & {76.65$\pm$1.21} & 73.60$\pm$1.25\\
        \hline
         NCA~\cite{wu2018improving} & \multirow{8}*{WRN-28-10} & 56.32$\pm$1.29 & 56.41$\pm$1.46 &  67.18$\pm$1.15 & 68.13$\pm$1.22 \\
        ~ Baseline~\cite{chen2019closer} & & 59.43$\pm$1.34 & 59.61$\pm$1.50 & 74.28$\pm$1.14 & 75.26$\pm$1.17 \\
        ~ S2M2\_R~\cite{mangla2020charting} & & 61.37$\pm$1.33 &  61.52$\pm$1.52& {79.83$\pm$1.34} & {80.69$\pm$1.36} \\
        ~ NegMargin~\cite{liu2020negative} & & 58.00$\pm$1.44 & 57.50$\pm$1.65 & 70.12$\pm$1.30 & 71.07$\pm$1.36 \\
        ~ \textcolor{mycolor}{PT+NCM~\cite{hu2021leveraging}} & & \textcolor{mycolor}{60.92$\pm$1.68} & \textcolor{mycolor}{61.12$\pm$1.73} & \textcolor{mycolor}{74.33$\pm$1.48} & \textcolor{mycolor}{74.96$\pm$1.51} \\
        ~ \textcolor{mycolor}{PEM$_{b}$E$\text{-}$NCM~\cite{hu2022squeezing}} & & \textcolor{mycolor}{60.40$\pm$1.72} & \textcolor{mycolor}{60.57$\pm$1.77} & \textcolor{mycolor}{72.63$\pm$1.48} & \textcolor{mycolor}{73.01$\pm$1.50} \\
        ~ \textcolor{mycolor}{EASY~\cite{bendou2022easy}} & & \textcolor{mycolor}{61.02$\pm$1.67} & \textcolor{mycolor}{61.25$\pm$1.71} & \textcolor{mycolor}{75.98$\pm$1.41} & \textcolor{mycolor}{76.43$\pm$1.43} \\
        ~ \textbf{SCAN (ours)} & & \textbf{66.75$\pm$1.35} & \textbf{67.71$\pm$1.45} & \textbf{82.57$\pm$1.13} & \textbf{83.73$\pm$1.12} \\

    \bottomrule
    \end{tabular}
    
    \label{table:Derm_similar}
\end{table*}

\begin{table*}
    \centering
    \caption{{Ablation studies on the improvements with various backbone architectures on the SD-198 dataset. The accuracy (\%) of 2-way and 5-way settings are displayed.} }
    \begin{tabular}{c|c|cccc}
    \toprule
    Method&Backbone&2-way 1-shot&2-way 5-shot&5-way 1-shot&5-way 5-shot\\
    \midrule
        
        Baseline \cite{chen2019closer} & \multirow{2}*{Conv4} & 73.98$\pm$1.38  & 88.67$\pm$1.01  & {80.62$\pm$0.41}  & {89.40$\pm$0.36}  \\
        ~ \textbf{SCAN (ours)} & & 77.12$\pm$1.44   & 90.22$\pm$0.95  & {81.85$\pm$0.41}  & {89.79$\pm$0.34}  \\
        \hline
         Baseline \cite{chen2019closer} & \multirow{2}*{Conv6} & 73.48$\pm$1.43  & 86.35$\pm$1.08  & {80.03$\pm$0.42}  & {87.63$\pm$0.37}  \\
        ~ \textbf{SCAN (ours)} & & 76.75$\pm$1.42   & 87.45$\pm$1.08  & {81.37$\pm$0.42}  & {89.53$\pm$0.36}  \\
        \hline
         Baseline \cite{chen2019closer} & \multirow{2}*{ResNet18} & 71.52$\pm$1.45  & 84.22$\pm$1.15 & {78.95$\pm$0.41}  & {86.87$\pm$0.37}  \\
        ~ \textbf{SCAN (ours)} & & 73.57$\pm$1.53   & 85.83$\pm$1.13  & {80.19$\pm$0.40}  & {88.30$\pm$0.34}  \\
        \hline
         Baseline \cite{chen2019closer} & \multirow{2}*{ResNet34} & 71.92$\pm$1.45  & 84.53$\pm$1.17  & {78.83$\pm$0.40}  & {86.80$\pm$0.37}  \\
        ~ \textbf{SCAN (ours)} & & 73.08$\pm$1.42   & 85.03$\pm$1.12  & {79.54$\pm$0.42}  & {87.45$\pm$0.40}  \\
        \hline
         Baseline \cite{chen2019closer} & \multirow{2}*{WRN-28-10} & 75.72$\pm$1.47  & 88.95$\pm$1.00  & {80.63$\pm$0.42}  & {89.46$\pm$0.38}  \\
        ~ \textbf{SCAN (ours)} & & 80.20$\pm$1.44   & 91.48$\pm$0.88  & {83.23$\pm$0.44}  & {92.16$\pm$0.31}  \\

    \bottomrule
    \end{tabular}
    \vspace{-2mm}
    \label{table:SD_backbones}
\end{table*}

To evaluate the performance on the novel set $D_n$, we freeze the parameters of the feature extractor $f_{\theta}$ and use it to encode images in a novel set to feature embeddings. Episodes are generated from the novel set randomly. We set 2-way 1-shot, 2-way 5-shot, 5-way 1-shot and 5-way 5-shot for the SD-198 dataset, and set 2-way 1-shot and 2-way 5-shot for the Derm7pt dataset. 5 query images are selected in each episode for both two datasets. For each episode, we train an episodic-specific linear classifier on the support set with the augmentation techniques introduced in \cite{yang2021free}, and test the average accuracy and the F1-score on the query set. The final results are reported as the mean classification accuracy over 600 randomly sampled episodes with 95 confidence intervals.

\if 1
\begin{table*} [!htbp]
    \centering
    \caption{Existing state-of-the-art FSL methods can be enhanced with our proposed SCA branch. Results (\%) on the SD-198 dataset {with 2-way 1-shot and 2-way 5-shot settings}. (Backbone: WRN-28-10).}
    \begin{tabular}{c|cccc}
    \toprule
    \multirow{2}{*}{Method} & \multicolumn{2}{c}{2-way 1-shot} & \multicolumn{2}{c}{2-way 5-shot}   \\
    & Accuracy & F1-score & Accuracy & F1-score \\
    \midrule
        
         S2M2\_R \cite{mangla2020charting} & 76.42$\pm$1.52 & 77.51$\pm$1.59 & 90.32$\pm$0.89 & 90.97$\pm$0.89  \\
        ~ \textbf{S2M2\_R+SCAN (ours)} & \textbf{79.23$\pm$1.41} & \textbf{80.27$\pm$1.44} & \textbf{91.00$\pm$0.90} & \textbf{91.60$\pm$0.88} \\
         \hline
         ~ NegMargin \cite{liu2020negative} & 76.85$\pm$1.39 & 77.98$\pm$1.45  & 89.92$\pm$0.96  &  90.65$\pm$0.92 \\
        ~ \textbf{NegMargin+SCAN (ours)} &\textbf{78.60$\pm$1.44} & \textbf{79.79$\pm$1.46} & \textbf{91.43$\pm$0.90} & \textbf{92.07$\pm$0.85} \\
    \bottomrule
    \end{tabular}
    
    \label{table:SD_further_2way}
\end{table*}
\begin{table*} [!htbp]
    \centering
    \caption{Existing state-of-the-art FSL methods can be enhanced with our proposed SCA branch. Results (\%) on the SD-198 dataset {with 5-way 1-shot and 5-way 5-shot settings}. (Backbone: WRN-28-10).}
    \begin{tabular}{c|cccccccc}
    \toprule
    \multirow{2}{*}{Method} & \multicolumn{4}{c}{5-way 1-shot} & \multicolumn{4}{c}{5-way 5-shot}   \\
    & Sensitivity & Specificity & Accuracy & F1-score & Sensitivity & Specificity & Accuracy & F1-score \\
    \midrule
         S2M2\_R \cite{mangla2020charting} & 54.79$\pm$1.07 & 88.70$\pm$0.27 & 81.91$\pm$0.43 & 55.49$\pm$1.13 & 77.11$\pm$0.83 & 94.28$\pm$0.21 & 90.84$\pm$0.33 & 78.17$\pm$0.84\\
        ~ \textbf{S2M2\_R+SCAN (ours)} & \textbf{57.35$\pm$1.07} & \textbf{89.34$\pm$0.27} &  \textbf{82.94$\pm$0.43} &
        \textbf{58.10$\pm$1.14} & \textbf{78.19$\pm$0.82} & \textbf{94.55$\pm$0.20} &  \textbf{91.28$\pm$0.33} & \textbf{79.58$\pm$0.80}\\
        \hline
        ~ NegMargin \cite{liu2020negative} & 55.38$\pm$1.07 & 88.34$\pm$0.27 &  82.15$\pm$0.43 & 56.04$\pm$1.14 & 76.71$\pm$0.87 & 94.18$\pm$0.22 & 90.68$\pm$0.35 & 77.75$\pm$0.87\\
        ~ \textbf{NegMargin+SCAN (ours)} &\textbf{59.18$\pm$1.08} & \textbf{89.80$\pm$0.27} &  \textbf{83.67$\pm$0.43} &
        \textbf{60.03$\pm$1.16} & \textbf{80.81$\pm$0.82} & \textbf{95.20$\pm$0.20} & \textbf{92.33$\pm$0.33} & \textbf{81.95$\pm$0.80} \\

    \bottomrule
    \end{tabular}
    
    \label{table:SD_further_5way}
\end{table*}
\fi

\subsection{Results}
\subsubsection{Results on SD-198 dataset}
The results of our proposed SCAN approach together with several other different methods on the SD-198 dataset are reported in Tables \ref{table:SD_similar_2way} and \ref{table:SD_similar_5way}. 
\tnnls{Table \ref{table:SD_similar_2way} shows the accuracy and F1-score with 2-way 1-shot and 2-way 5-shot settings, and Table \ref{table:SD_similar_5way} lists the sensitivity, specificity, accuracy and F1-score with 5-way 1-shot and 5-way 5-shot settings.}
Among the compared methods, PCN \cite{prabhu2019few} and Meta-derm \cite{mahajan2020meta} are two meta-learning-based FSL algorithms designed for rare skin disease classification. 
\tnnls{PCN~\cite{prabhu2019few} uses the sub-cluster idea which is similar to ours. However, it divides each class into the same number of sub-clusters, which fails to represent the unique structures of different classes. When we consider the diverse numbers of sub-clusters for various classes, the model learns more accurate feature embeddings for the base set and performs better results on the novel set.} 
Meta-derm \cite{mahajan2020meta} is built on Reptile and ProtoNets by using a substitute Group Equivariant Convolutions, and \revise{they} present the \revise{state-of-the-art} accuracy on the SD-198 dataset \revise{with 2-way 1-shot and 5-shot settings}. They do not consider the large intra-class variation issue for skin diseases, therefore our method exceeds the accuracy of the Meta-derm method by 11.45\% and 3.75\% in 2-way 1-shot and 2-way 5-shot experiments, respectively. 

Apart from the meta-learning-based methods, we also conduct comparison experiments on seven state-of-the-art transfer-learning-based FSL methods. 
NCA~\cite{wu2018improving} proposes neighbor component analysis loss which can learn the relationships between features. The Baseline~\cite{chen2019closer} method was proposed as the baseline model for transfer-learning-based FSL methods. It trains the feature encoder with a standard cross-entropy loss. S2M2\_R~\cite{mangla2020charting} utilizes self-supervision and Manifold Mixup to enhance the feature encoder. NegMargin~\cite{liu2020negative} applies negative margin softmax loss when training on the base set and obtains increased results on the novel set. \tnnls{Instead of training a powerful feature encoder in the pre-training step, some methods apply feature post-processing tricks in the fine-tuning step. PT+NCM~\cite{hu2021leveraging} proposes Power Transform (PT) on the support and the query features to align their distributions closer to Gaussian-like distributions. In addition to PT, PEM$_{b}$E~\cite{hu2022squeezing} applies Euclidean normalization and Mean subtraction to further reduce the task bias. EASY~\cite{bendou2022easy} uses random resized crops to augment support set images and leverages the ensemble tricks to further boost the performance.
Compared to them, SCAN pays more attention to the feature encoder training instead of the post-processing tricks. We propose the specific unsupervised cluster branch to handle the inherent sub-cluster issue in skin disease datasets. Therefore, our model learns more precise feature embeddings on the base set and shows better performance on the novel set.}

\begin{table*}
    \centering
    \caption{\tnnls{Ablation studies on the effectiveness of cluster branch and the purity loss on the SD-198 dataset. The accuracy (\%) of 2-way and 5-way settings and the cluster error rate (\%) are reported.}}
    \begin{tabular}{ccc|ccccc}
    \toprule
    {\thead{Class branch \\ ($\mathcal{L}_{class}$) }} & {\thead{Cluster branch \\ (+$\mathcal{L}_{cluster}$)}} & {\thead{Cluster branch \\ (+$\mathcal{L}_{purity}$)}} &2-way 1-shot&2-way 5-shot&5-way 1-shot&5-way 5-shot& Cluster Error Rate\\
    \midrule
        \checkmark & & & 75.72$\pm$1.47 & 88.95$\pm$1.00 & {80.63$\pm$0.42} & {89.46$\pm$0.38} & - \\
         & \textcolor{mycolor}{\checkmark} & & \textcolor{mycolor}{63.37$\pm$1.46} & \textcolor{mycolor}{75.28$\pm$1.25} & \textcolor{mycolor}{71.58$\pm$0.42} & \textcolor{mycolor}{78.54$\pm$0.39} & \textcolor{mycolor}{61.03} \\
        \checkmark & \checkmark & & 78.65$\pm$1.46 & 89.46$\pm$0.98 & {81.29$\pm$0.42} & {90.88$\pm$0.37} & 35.10 \\
        \checkmark & \checkmark & \checkmark & \textbf{80.20$\pm$1.44} & \textbf{91.48$\pm$0.88} & {\textbf{83.23$\pm$0.44}} & {\textbf{92.16$\pm$0.31}} & \textbf{20.58} \\

    \bottomrule
    \end{tabular}
    \vspace{-2mm}
    \label{table:SD_purity_loss}
\end{table*}
\begin{figure*}
\centering
\includegraphics[width=1.0 \textwidth]{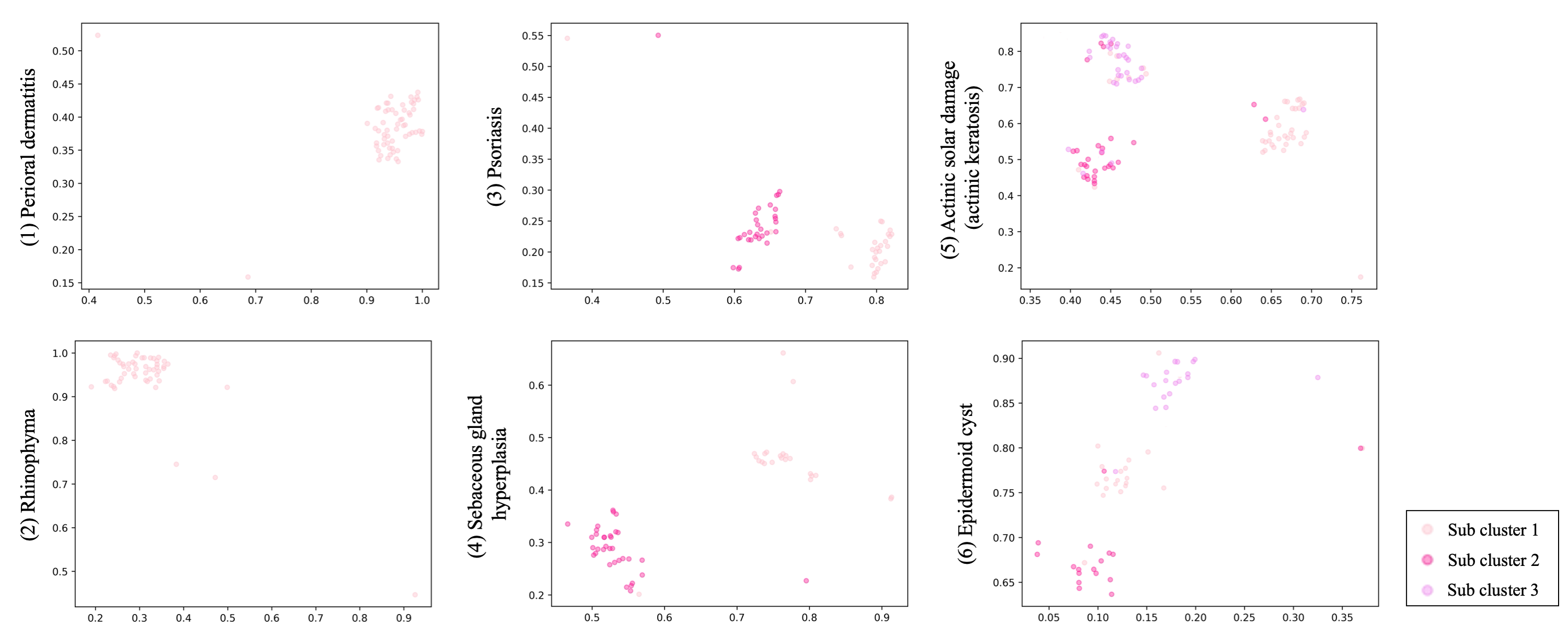} 
\caption{\tnnls{The t-SNE visualization results of feature embeddings learned by SCAN. Six classes in the base set of the SD-198 dataset are shown in (1) – (6). Various numbers of sub-clusters are identified for different classes.}} 
\vspace{-4mm}
\label{fig7.0} 
\end{figure*}

\subsubsection{Results on Derm7pt dataset}
The results of our proposed SCAN method and other compared methods on the Derm7pt dataset are presented in Table~\ref{table:Derm_similar}. We compare SCAN with two state-of-the-art meta-learning and seven transfer-learning FSL methods. In general, the experimental results show that our method provides superior performance on the Derm7pt dataset among all the compared algorithms. For the setting of 2-way 5-shot, SCAN exceeds all the compared methods except for Meta-derm~\cite{mahajan2020meta} when using Conv6 as the backbone architecture. 

\if 1
\subsubsection{Plug-and-Play module}
Our SCAN framework can be used as a plug-and-play branch to enhance existing FSL methods. We validate the performance by combining \revise{the cluster branch of SCAN} with two state-of-the-art FSL algorithms: S2M2\_R \cite{mangla2020charting} and NegMargin \cite{liu2020negative}. They both follow the transfer-learning training paradigm.  
To combine our proposed cluster branch with these existing approaches, we append the projection head $\mathbf{z}_{\theta}$ and the cluster classifier $C(.|\mathbf{W}_{cluster}^{base})$ to the penultimate layer of the original S2M2\_R and NegMargin frameworks. Also, two losses $\mathcal{L}_{cluster}$ and $\mathcal{L}_{purity}$ are added to the original training loss of the two methods. The experimental results on the SD-198 dataset are summarized in Tables \ref{table:SD_further_2way} and \ref{table:SD_further_5way}. We find that our proposed cluster branch boosts the accuracy of S2M2\_R \cite{mangla2020charting} and NegMargin \cite{liu2020negative} methods by around 1\%-3\%.

\fi


\subsection{Ablation Study}
\subsubsection{Backbone architectures}
In addition to the WRN-28-10 architecture, we conducted performance tests of the SCAN method on Conv4, Conv6, ResNet18, and ResNet34 architectures to assess its robustness across different backbone networks. We compared SCAN with the Baseline method \cite{chen2019closer}, which utilizes only the $\mathcal{L}_{class}$ loss. The results are presented in Table \ref{table:SD_backbones}.
Our proposed method, SCAN, consistently improves performance across various backbone networks. However, we observed that the performance of ResNet backbones generally falls behind that of Conv backbones. This discrepancy may be attributed to overfitting on the base set. By increasing the width of the residual blocks, the overfitting problem is mitigated. The best performance is achieved when employing the WideResNet (WRN) backbone.

\subsubsection{Effectiveness of each component}
\tnnls{We conducted evaluations to assess the effectiveness of each component in the SCAN framework. The results are presented in Table~\ref{table:SD_purity_loss} and include four scenarios: 1) using only the class branch, 2) using only the cluster branch, 3) using both the class and cluster branches without the purity loss, and 4) using both the class and cluster branches with the purity loss. In addition to reporting the accuracy of the novel set, we introduced a metric called the \textit{cluster error rate} to evaluate the quality of the learned sub-clusters on the base set. The cluster error rate measures the percentage of samples that are assigned to incorrect clusters out of the entire dataset. It is calculated by summing the number of items with different class labels from the major examples for each cluster and dividing it by the total size of the training set. 

Based on the results in Table~\ref{table:SD_purity_loss}, we highlight the following three key findings. Firstly, models trained with only the class branch or the cluster branch exhibit poor performance, \tnnlsminor{with 2-way 1-shot accuracy of 75.72\% and 63.37\%, respectively.} The cluster branch performs worse than the class branch, likely due to the absence of supervised class labels in the cluster branch. Secondly, combining both the class and cluster branches leads to a significant improvement in accuracy. \tnnlsminor{For instance, the accuracy of the 2-way 1-shot setting is enhanced from 75.72\% to 78.65\%.} This indicates the effectiveness of addressing the challenge posed by large intra-class variation. Thirdly, the inclusion of the purity loss helps refine the clustering results, resulting in a significant decrease in the cluster error rate. This demonstrates the ability of the purity loss to enhance the quality of the learned sub-clusters.
Overall, these findings confirm the effectiveness of the different components in the SCAN framework for learning dynamic sub-cluster structures and improving classification performance.


} 


\begin{figure*}
\centering
\includegraphics[width=0.8 \textwidth]{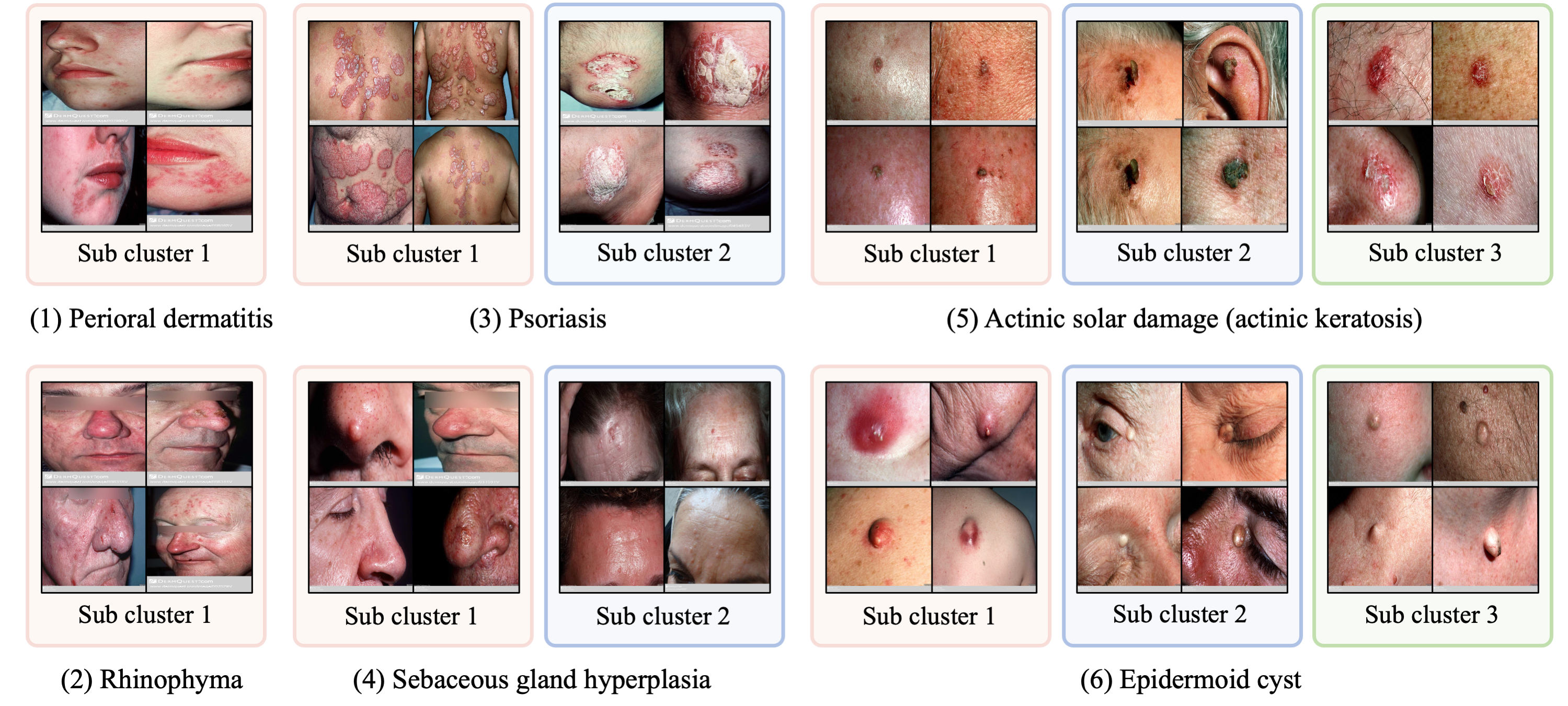} 
\caption{\tnnls{Examples of sub-clusters learned by SCAN. Images from six classes in the base set of the SD-198 dataset are shown in (1) – (6).}} 
\label{fig7} 
\end{figure*}
\begin{table*} [t]
    \centering
    \caption{Discriminative analysis on the base and novel classes of SD-198 dataset. The inter-class distance ($D_{inter}$), the intra-class distance ($D_{intra}$), and the discriminative index ($\phi$) are computed.}
     
    \begin{tabular}{cc|ccc|ccc}
    
    \toprule
     \multirow{2}{*}{Class branch} & \multirow{2}{*}{Cluster branch} & \multicolumn{3}{c|}{Base classes} & \multicolumn{3}{c}{Novel classes}   \\
     &  & $D_{inter}$ & $D_{intra}$ & $\phi$ & $D_{inter}$ & $D_{intra}$ & $\phi$ \\

    \midrule
        
          \checkmark & & 1.7251 & 0.1801 & 9.5777 & 0.7631 &  0.6235 & 1.2283 \\
           & \checkmark & 0.5990 & 0.6964 & 0.8601 & 0.7807 &  0.6739 & 1.1585  \\
           \checkmark & \checkmark & 1.6377 & 0.5067 & 3.2321 & 0.8318 &  0.5893 & 1.4115  \\

    \bottomrule
    \end{tabular}
    \vspace{-2mm}
    \label{table:disc_analysis}
\end{table*}

\subsubsection{Visualization of feature distribution in base classes}
To showcase the capability of our proposed SCAN framework in learning the dynamic sub-cluster structures, we utilize t-SNE visualization~\cite{van2008visualizing} to depict the feature distributions within the base classes of the SD-198 dataset. Figure~\ref{fig7.0} presents the t-SNE visualization of six base classes, where the pseudo sub-cluster labels learned by SCAN are leveraged to illustrate the feature distribution within each class. The sub-cluster structures exhibited in Figure~\ref{fig7.0} reveal varying numbers of sub-clusters for each class, ranging from one to three. This dynamic identification of sub-cluster numbers for different classes demonstrates the capability of SCAN.

To validate the rationality of the clustering outcomes, we display some representative images from each sub-cluster for the six base classes in Figure~\ref{fig7} (1) -- (6). For the classes perioral dermatitis and rhinophyma, the entire class is considered a unified group due to the high similarity among items. As depicted in Figure~\ref{fig7} (1) and (2), the affected regions of these two classes are mostly concentrated in the same area (the chin region for perioral dermatitis and the nose area for rhinophyma). In cases where a class forms multiple sub-clusters, the images within different sub-clusters originate from distinct body locations, resulting in diverse symptoms. 

In summary, our proposed method effectively learns the cluster arrangement within each class through the incorporation of an additional unsupervised cluster branch and two restricted losses. This enables the dynamic exploration of sub-cluster structures within the dataset.

\subsubsection{Discriminability analysis on feature embeddings}
In this section, we explore the impact of the proposed method on the features in the base and novel set. In the pre-training stage on the base classes, the class branch aims to increase the distance among different class centers (\ie, the inter-class distance), while the cluster branch aims to increase the distances within a class (\ie, the intra-class distance). 

To quantitatively measure the impact of two branches on both the base and novel classes, we compute the inter-class distance ($D_{inter}$), the intra-class distance ($D_{intra}$), and the discriminative index ($\phi$), based on the definitions provided in~\cite{liu2020negative,mika1999fisher}. The equations of three metrics are:
\begin{equation}
D_{inter} = \frac{1}{k(k-1)}\sum_{m=1}^{k}\sum_{n=1, n\neq m}^{k} \|\mu(c_m)-\mu(c_n)\|_2^2,\label{inter_variance}
\end{equation}
\begin{equation}
D_{intra} = \frac{1}{k}\sum_{m=1}^{k}(\frac{1}{|{c_m}|}\sum_{(x_i,y_i)\in c_m}^{k} \|\frac{f_{\theta}(x_i)}{\|f_{\theta}(x_i)\|}-\mu(c_m)\|_2^2),\label{intra_variance}
\end{equation}
\begin{equation}
\phi = \frac{D_{inter}}{D_{intra}},\label{phi}
\end{equation}
where $\mu(c_m)$ denotes the center of class $c_m$ and is calculated by the mean of the L2-normalized feature embeddings of class $c_m$. 

$D_{inter}$ is computed by the mean L2 distances between every pair of class centers. $D_{intra}$ is computed by the mean L2 distances between every sample in a class and its corresponding class center. The discriminative index $\phi$ is defined as the division of $D_{inter}$ and $D_{intra}$. The higher value of $\phi$ indicates more discriminative feature embeddings, as it indicates a larger ratio of inter-class distance and intra-class distance.

\begin{figure*}
\centering
\includegraphics[width=1.0\textwidth]{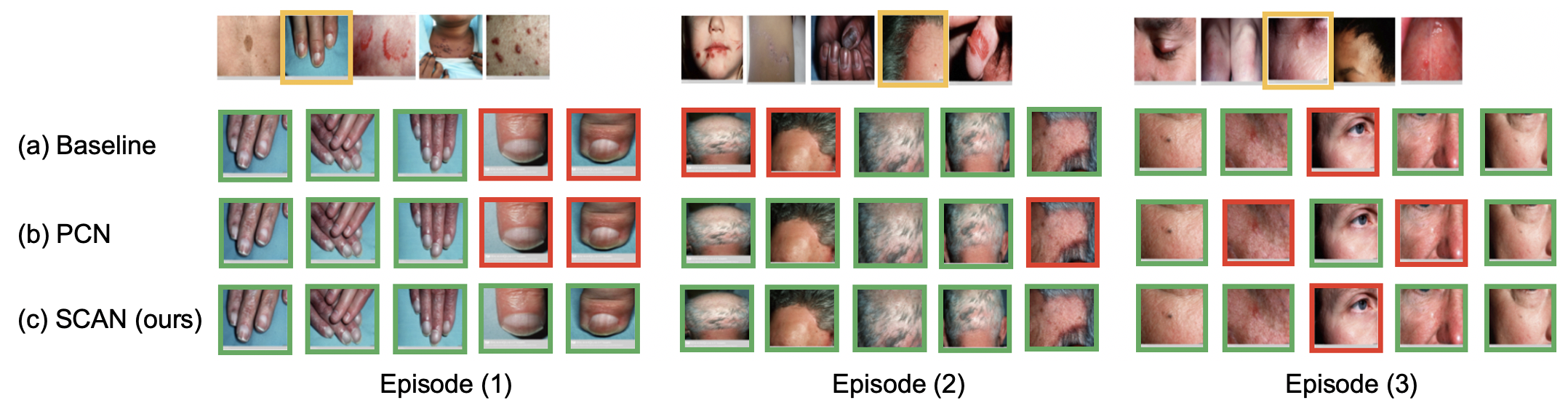} 
\caption{Results of query images prediction on the SD-198 novel set for 5-way 1-shot task. (1) (2) (3) are sampled episodes. \revise{For each episode, in the top row are images from the support set. There are 5 classes in total and 1 image from each class. The remaining 5 images are the query images from one of the above 5 classes in the support set, and we use a yellow bounding box to show the target class. Three rows show the testing results of three methods:} (a) Baseline \cite{chen2019closer}, (b) PCN \cite{prabhu2019few} and (c) SCAN (ours). The green/red bounding boxes denote the true/false classification. } 
\label{Fig.classification_result} 
\end{figure*}
\begin{figure}
\centering
\includegraphics[width=0.49 \textwidth]{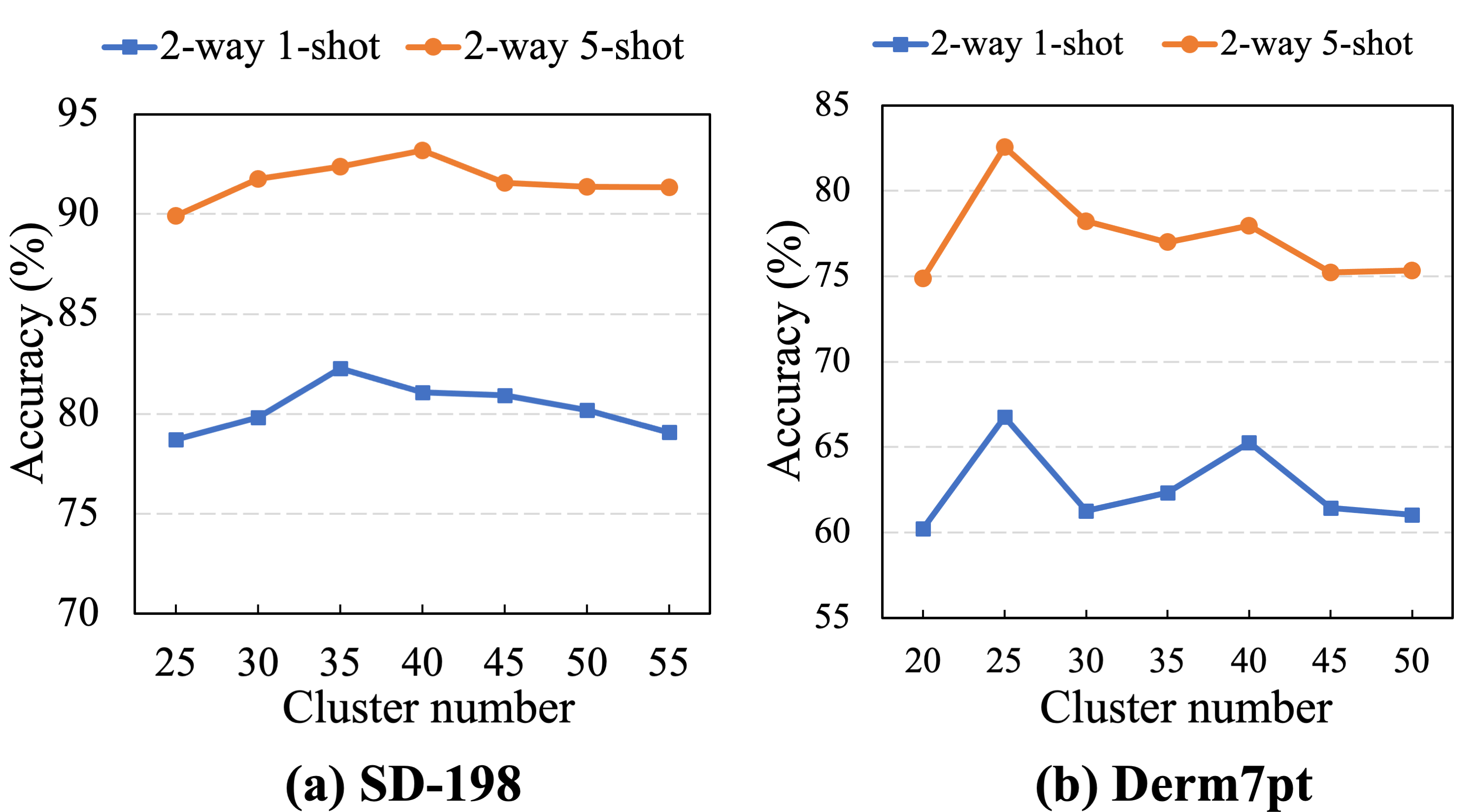} 
\caption{{Ablation studies on the hyperparameter in SCAN. The accuracy of 2-way 1-shot and 2-way 5-shot is plotted with respect to the cluster number for both the SD-198 and Derm7pt datasets.}} 
\vspace{-4mm}
\label{fig7.1} 
\end{figure}

\tnnls{We present the results of $D_{inter}$, $D_{intra}$, and $\phi$ on the base and novel classes of the SD-198 dataset in Table~\ref{table:disc_analysis}. For the base classes, we observe that the class branch increases the inter-class distance from 0.5990 to 1.6377, and the cluster branch increases the intra-class distance from 0.1801 to 0.5067. These results validate the roles of two branches on the base classes, as mentioned in our previous intuitive explanation. As a result of the increased intra-class distance and relatively unchanged inter-class distance, the discriminative index decreases for the base classes when adding the cluster branch to the class branch. However, we argue that less discrimination in base set features does not indicate a conflict between two branches, as our primary focus is the performance of the novel classes. For the novel classes, the combination of two branches leads to an increase in inter-class distance and a decrease in intra-class distance compared to using either the class or the cluster branch alone. 

These results demonstrate that our proposed method enables the feature encoder to embed more generalizable feature representations by learning the sub-cluster structures of the base classes. Consequently, the feature encoder can extract more discriminative features for the novel classes, resulting in the improved classification accuracy.}

\subsubsection{Visualization of classification results on novel classes}
To compare the episodic predictions among various methods, we display three examples for the 5-way, 1-shot experiment of (a) Baseline \cite{chen2019closer}, (b) PCN \cite{prabhu2019few} and (c) our proposed SCAN method, as depicted in Figure \ref{Fig.classification_result}. The first row shows the sampled support set images, with one example per class from a total of five classes. The yellow bounding box indicates the ground truth category for the query images.
The subsequent three rows illustrate the results of the query images for each method. The green and red boxes represent correct and incorrect predictions, respectively, on the target class in the support set.

In general, our method demonstrates higher accuracy in predicting the listed query set compared to the other methods, particularly evident in episodes (1) and (2) of Figure~\ref{Fig.classification_result}. While our method correctly identifies examples that exhibit high similarity to the support image, such as the first three images in (1), it also effectively handles cases that are distinguishable from the given support item. For instance, it accurately predicts the last two columns in (1) and the first and fifth columns in (2).
However, it is worth noting that the incorrect prediction in the third column of episode (3) suggests that the performance may be influenced by large inter-class similarity. This is evident as the third query image bears more resemblance to the support image from the first class than to the image sampled from its own class.


\begin{table}
    \centering
    \caption{The computational efficiency analysis of the proposed method.}
    \begin{tabular}{c|ccc}
    \toprule
    Method& Time per epoch &FLOPs&Model parameters\\
    \midrule
        Baseline~\cite{chen2019closer} & 12.28 sec & 32.81 GMac & 36.60 M \\
        PCN~\cite{prabhu2019few} & 16.44 sec & 32.81 GMac & 36.60 M \\
        SCAN (ours) & 20.93 sec & 32.81 GMac & 37.12 M \\
    \bottomrule
    \end{tabular}
    \vspace{-2mm}
    \label{table:computation}
\end{table}

\subsubsection{Ablation studies on hyperparameters}
We conducted additional experiments to assess the efficacy of the hyperparameters in SCAN. Figure~\ref{fig7.1} presents the results of 2-way 1-shot and 2-way 5-shot on two datasets, demonstrating their correlation with the cluster number. Our findings indicate that our method enhances overall performance across a range of cluster numbers. Notably, cluster numbers that are either too small or too large fail to adequately represent the sub-cluster structures. The highest accuracy is attained when the cluster number is 40 for the SD-198 dataset and 25 for the Derm7pt dataset.

\tnnls{
\subsubsection{Computational efficiency analysis}
To evaluate the computational efficiency of our proposed method, we conducted a comparison with the Baseline~\cite{chen2019closer} and PCN~\cite{prabhu2019few} methods on the SD-198-20 dataset. The Baseline and PCN methods have identical network architectures, which only contain the class branch. We utilized the same backbone, hyperparameters, and GPU for all methods. The results of time, floating-point operations (FLOPs), and model parameters are reported in Table~\ref{table:computation}. 
We trained all three methods for 800 epochs and recorded the average time consumed for one epoch. Our proposed SCAN method takes approximately 8 seconds longer than the Baseline method and 4 seconds longer than the PCN method to complete the clustering calculations. However, our performance is vastly superior to the Baseline and PCN methods, improving the accuracy by 4.48\% and 7.09\% respectively.
We also computed the number of FLOPs for all methods, using an input size of 3x80x80 and a batch size of 1. The results show that the FLOPs numbers are nearly the same for all three methods. The additional computational effort of the linear classifier in the cluster branch of SCAN can be considered negligible.
Lastly, we compared the number of parameters of SCAN with the other two methods. The SCAN method has approximately 0.52M more parameters than the Baseline and PCN methods. These additional parameters in SCAN are derived from the extra linear classifier in the cluster branch. 

Overall, our proposed method achieves superior performance compared to the Baseline and PCN methods, while the additional computational cost is minimal. This highlights the effectiveness and efficiency of our proposed method for rare skin disease diagnosis.}

\section{Conclusion}


In this paper, we present a novel Sub-Cluster Aware Network (SCAN) for rare skin disease classification. According to our key insights, skin disease datasets have existing latent sub-groups within a class. Therefore, our proposed SCAN effectively learns the intrinsic sub-cluster structures of skin disease via a well-designed dual branch framework and three additive losses. The results tested on two public skin image datasets show that our method excels over other state-of-the-art methods by around 2\% to 5\% for various settings.


\bibliographystyle{IEEEtranN}
\small{\bibliography{refs}}
\end{document}